\documentclass[sigconf]{acmart}
\usepackage{multirow}
\usepackage[textsize=scriptsize]{todonotes}
\pdfoutput=1
\usepackage{booktabs} % For formal tables
\renewcommand\footnotetextcopyrightpermission[1]{} 

\setcopyright{rightsretained}
\acmDOI{10.475/123_4}

\acmISBN{123-4567-24-567/08/06}

\acmConference[]{CODS-COMAD}{Jan 2019}{Kolkata, WB, India}
\acmYear{1997}
\copyrightyear{2016}

\acmArticle{4}
\acmPrice{15.00}

\begin{document}
\title{CQASUMM: Building References for Community Question Answering Summarization Corpora}
\titlenote{The work was done when the first author was at IIIT Delhi, India.}

\author{Tanya Chowdhury}
\affiliation{\institution{Myntra Designs} \city{Bengaluru}
  \state{India}}

\email{tanya.chowdhury@myntra.com}

\author{Tanmoy Chakraborty}
\affiliation{\institution{IIIT-Delhi} \city{New Delhi}
  \state{India}}
\email{tanmoy@iiitd.ac.in}

% The default list of authors is too long for headers.
\renewcommand{\shortauthors}{Tanya Chowdhury, Tanmoy Chakraborty}

\begin{abstract}
Community Question Answering forums such as Quora, Stackoverflow are rich knowledge resources, often catering to information on topics overlooked by major search engines. 
Answers submitted to these forums  are often elaborated, contain spam, are marred by slurs and business promotions. It is difficult for a reader to go through numerous such answers to gauge community opinion. As a result summarization becomes a prioritized task for CQA forums. While a number of efforts have been made to summarize factoid CQA, little work exists in summarizing non-factoid CQA. We believe this is due to the lack of a considerably large, annotated dataset for CQA summarization. We create {\bf CQASUMM}, the first huge annotated CQA summarization dataset by filtering the 4.4 million Yahoo! Answers L6 dataset. We sample threads where the best answer can double up as a reference summary and build hundred word summaries from them.  We treat other answers as candidates documents for summarization. We provide a script to generate the dataset and introduce the new task of {\em Community Question Answering Summarization}.  

Multi document summarization has been widely studied with news article datasets, especially in the DUC and TAC challenges using news corpora. However documents in CQA have higher variance, contradicting opinion and lesser amount of overlap. We compare the popular multi document summarization techniques and evaluate their performance on our CQA corpora. We look into the state-of-the-art and understand the cases where existing multi document summarizers (MDS) fail. We find that most MDS workflows are built for the entirely factual news corpora, whereas our corpus has a fair share of  opinion based instances too. We therefore introduce {\bf OpinioSumm}, a new MDS which outperforms the best baseline by $4.6\%$ w.r.t ROUGE-1 score. 
 To increase reproducibility of our work, we make the code and dataset public at \url{https://bitbucket.org/tanya14109/cqasumm}.
\end{abstract}

\if 0
%
% The code below should be generated by the tool at
% http://dl.acm.org/ccs.cfm
% Please copy and paste the code instead of the example below.
%



\fi

\maketitle

\vspace{-5mm}

\section{Introduction}

%What are CQAs?
Community Question Answering (CQA) are websites known to archive millions of question-answer pairs contributed by community users. They form a rich knowledge base, often missing from web search engines. Most of these forums are unmoderated and open to join and post. Some of them may allow virtual anonymity. 
Of late, community question answering services like Quora\footnote{https://www.quora.com/} have gained a lot of popularity. Services such as Stackoverflow\footnote{https://stackoverflow.com/}, Mathexchange\footnote{https://math.stackexchange.com/} etc. cater to a specific scientific community. 
%There also exist product dependent consumer forums like askubuntu etc.
With the advent of so many forums, tasks related to them also gain importance. Some widely studied problems are finding similar questions, suggesting most relevant answer, ranking of answers, etc.
%Need for summarization in CQAs
Some of these forums are unstructured and have hundreds of answers corresponding to a particular question. It is difficult for a user looking for an answer to gauge community opinion by going through each answer. These forums may also be used for self advertisement or to spread propaganda. It is thus an important task to summarize these information rich resources. 

%Quora answer wiki and what do you mean by summary
Throughout this work, we refer to a CQA question and its set of answers as a question thread. Answers belonging to a CQA thread are referred to as candidate documents. A few years ago, Quora introduced the concept of Answer Wikis\footnote{https://www.quora.com/What-is-an-Answer-Wiki-on-Quora-1} as a manual, collaborative aggregated answer for a question thread. They started with it to provide an impartial articulation of the leading perspectives in the question thread, a collection of factual and uncontroversial opinion. This is the definition of a question thread summary we keep in mind, during the entirety of this work. Answer Wikis are user-editable and are usually found on the top of the question page. They act as a `TL;DR' (too long; don't read) to the question thread, and edits to them require moderation by Quora. We introduce a task to auto generate such Answer Wikis. 

%Factoid vs non factoid summaries
A factoid question is a query which can be answered with concise facts. Non-factoid Q/A is an umbrella term which covers all topics beyond factoid Q/A\footnote{https://www.quora.com/Natural-Language-Processing-What-is-Non-factoid-question-answering}. Many works have studied how to generate summaries for factoid questions in CQA services, one of the most recent ones being \cite{chowdhury2018viz}. However little effort has been made towards generating summaries for non-factoid threads. This can be attributed to the open domain nature of questions in non-factoid Q/A resulting in a high variance in question and answer quality. This also makes CQA summarization a different problem from generic multi document summarization on a news article corpus, a problem which has been studied relatively widely. Another reason for the absence of work in this domain is the lack of a large enough CQA summarisation dataset with annotated summaries. Authors in the existing CQA summarization \cite{liu2008understanding,cao2011question} efforts have worked on small scale data which they have manually annotated and not made public. 

%Previous efforts of summarization in CQA. 
Following efforts have been made in the past for CQA summarization. \citet{liu2008understanding} and \citet{cao2011question} suggested methods depending on the taxonomy of the question. The former divided questions into one of Navigational, Informational, Transactional or Social. \citet{pande2013summarizing} picked an incomplete best answer and built a summary by adding valuable missing information to it. \citet{song2017summarizing} proposed a sparse coding based summarization strategy with short document expansion and sentence vectorization. 

Most of the above stated methods are extractive summarization, i.e., they pick important sentences from the answers. We again attribute the absence of abstractive summarization methods in CQA to the absence of a relevant annotated dataset required to train deep learning models. News corpora such as DUC\footnote{http://duc.nist.gov/} 2002 and 2004, TAC\footnote{http://tac.nist.gov/} or the CNN/Dailymail dataset are not a good fit for CQA training as repeated bigram frequency in different versions of a news article are higher than in CQA, where answers have a huge variance \cite{liu2008understanding}.
 In this work, we introduce a dataset of more than 1 million question threads which can be used to train/test CQA summarization algorithms.  
%Contributions
The contributions of this paper are as follows:
\begin{enumerate}
\item We provide a script to generate an annotated CQA Summarization dataset, called CQASUMM from the available Yahoo! Answers L6 dataset. We then introduce a new CQA summarization task. 
\item We compare popular multi document summarization techniques, previously tested on news document corpora, and evaluate their performance on CQASUMM. 
\item We introduce OpinioSumm, our proposed CQA summarization algorithm built on top of TextRank, which performs better than the best baseline by $4.6\%$  on question threads in the CQA corpus w.r.t ROUGE-1 score. 
\end{enumerate}

\if 0
The rest of the paper is ordered as follows. In Section 2, we do a survey of four categories, most multi document summarization techniques can be classified into. In Section 3 we explain our corpus, how it can be reconstructed and describe the task that this paper introduces. In Section 4, we introduce CQASumm, our CQA summarization algorithm based on sentence compression and fusion.  In Section 5 we discuss the quality of our corpus and evaluate and compare the performance of various multi document summarization techniques on our corpus. The later sections include related work, theory related to principles used, corpus reconstruction guide and a discussion followed by future work.  
\fi

\begin{figure}
\begin{center}

\includegraphics[height=4in, width=\linewidth]{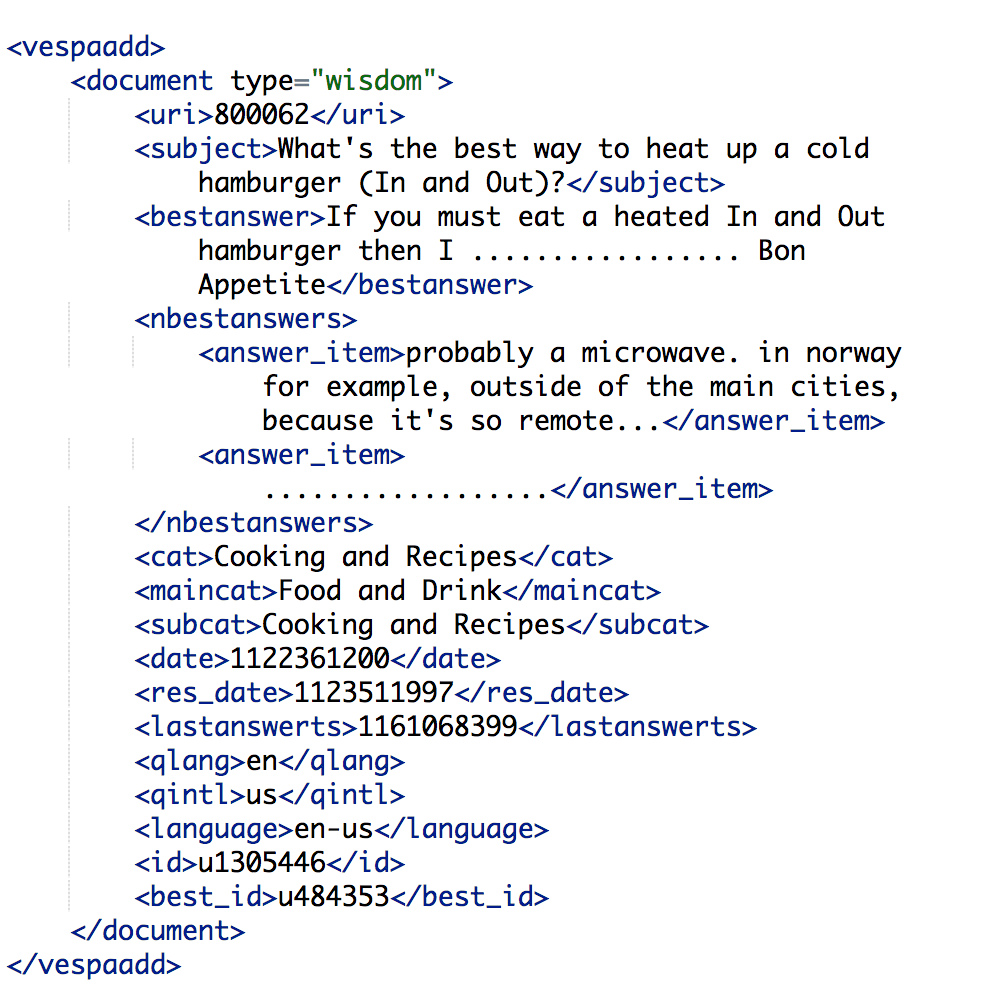}
\caption{A sample question thread from Yahoo! Answers corpus. Each question thread mandatorily has a main category, sub category, subject, best answer and list of other answers. It also has anonymized author ids, question language and question date. XML printed using \url{https://jsonformatter.org/xml-pretty-print}} \label{fig:xml}
\end{center}
\vspace{-5mm}
\end{figure}

\section{Background}
Summarization can be broadly classified as extractive and abstractive. In extractive methods, important sentences/phrases from the text are selected and concatenated to form a summary. In abstractive methods, the important sentences/phrases are restructured before concatenation. 
Before the advent of deep learning techniques, the DUC and TAC tasks led to a number of extractive summarization techniques, the quality of which was measured by computing ROUGE score with manually annotated summaries provided in the tasks. The tasks further had sub-tasks, majorly single document and multi document summarization.  

The multi document summarization problem differs from its single document counterpart as it may contain documents which overlap, supplement or contradict each other. The task remains to highlight popular opinion and produces a complete, non-redundant and coherent summary.   
Since the CQA summarization problem can be closely classified as the well-studied multi document summarization problem, we make a survey of State-of-the-art extractive multi document summarization techniques. 

The main ideas behind multi document summarization techniques can be divided into cluster based, knowledge based, graph based and feature based as discussed by \citet{Kumar12automaticmulti}.  
We roughly go through each of them:

\begin{itemize}
\item \textbf{Feature Based Techniques}: In this method we assign each sentence a weight based on certain features and the highest weighted sentences are concatenated to form a summary. Some features used to assign weights to a sentence are: constituent word frequency (calculated via tf-idf scores), title words (occurrence of words appearing in the title to make the sentence highly relevant), location (if the sentence appears in the beginning, middle or end of the document; sentences in the beginning of a paragraph/document are usually more important), sentence length (very short or very long sentences do not contain meaningful content for summary), cue word occurrence (words such as ``Importantly'', ``Significantly'' or ``In conclusion'' alleviate the importance of a sentence), proper noun occurrence (one of the most important features) and common noun occurrence.
Each sentence is scored according to each of the above metrics, and a linear combination of these metric scores is taken to assign a weight to a sentence. 
The weight for each feature can be learned during training as in \cite{binwahlan2009swarm} or determined using genetic algorithms as in \cite{bossard2011combining}

\item \textbf{Cluster Based Techniques}:
Sentences/paragraphs representing similar views are grouped together to form clusters. Sentences are usually represented as tf-idf vectors of constituent words; and the similarity of two sentences is calculated as the cosine similarity between two tf-idf vectors. The clustering can be agglomerative or divisive. Larger the size of a cluster, more popular the opinion.  In \cite{erkan2004lexpagerank} , sentences nearest to cluster centroids were picked to represent the cluster. Clusters are sorted in descending order in accordance to their sizes, and the centre vectors are concatenated until the maximum  upper word bound of the summary is reached. Clustering may be done via K-means. A task remains to identify the number of initial clusters for the data that needs to be fed to the K-means algorithm. \citet{xia2011co} used sentence-term co-occurrence matrix based method to determine optimal number of clusters. Overall clustering methods are good in accommodating diverse information; however the summaries formed are not coherent and lack contextual information. 

\item \textbf{Graph Based Techniques}: The documents are represented as fully connected graphs. Sentences are  vertices in the graph, and the edge weight between any two vertices represents the similarity between the two sentences. Cosine similarity is often the similarity measuring metric. An edge weight threshold is defined below which the edges are dropped. A sentence is considered as an important candidate for summary if it is strongly connected to neighboring sentences. The method in \cite{erkan2004lexrank} is based on a similar graph based ranking algorithm. In further improvisations of the method, inter document and intra document links may be given dissimilar weights. \citet{wan2006improved} assigned higher priorities to sentences with stronger inter document linking. Another modification includes removing an already selected sentence vertex from the graph before choosing following sentences \cite{hariharan2009studies}.

\item \textbf{Knowledge Based Techniques} : Another approach used for Multi document summarization is the use of background knowledge to understand a document. Documents in MDS are over a few topics, each of which may have their existing knowledge graph. These ontology driven models have been used to understand the semantic relation between different entities in different documents. Several CQA summarization approaches have used existing knowledge bases to gain more insight about the context.
\end{itemize}

Multi document summarization has been mostly performed on the news corpora, which is entirely factual information. It has sometimes been performed on the user reviews corpora, which is entirely opinion based. The methods employed in the above two summarization workflows differ substantially. Our CQA corpora however has almost equal instances of factual and opinion based snippets. We believe that we can do better than a generic algorithm in order to increase ROUGE scores in such mixed corpora. We thus present OpinioSumm, our CQA summarization method.

\begin{figure*}
\begin{center}
\includegraphics[height=2in, width=\linewidth]{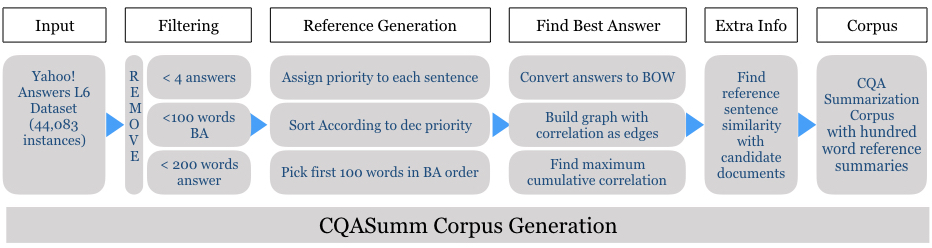}
\caption{We execute the following steps to generate CQASumm corpus. We begin with the Yahoo! Answers L6 Dataset which has 44,083,000 instances. We filter out threads with less than 4 answers, best answers less than 100 words etc. We then compress the best answer to generate a 100 word reference summary. We validate our reference summary by determining if the annotated best answer is indeed the unique best answer to the question thread. Lastly we filter out threads where majority of the views in the reference summary are absent from other answers.} \label{fig:CQASummPipeline}
\end{center}

\end{figure*} 

\section{Corpus Construction}
We begin with the L6 - Yahoo! Answers Comprehensive Questions and Answers dataset. Given the unstructured nature of text and high amount of sarcastic answers in the corpus, it would have been a better choice to use newer moderated platforms. Quora, for example, does not allow submission of answers below a certain word limit or plagiarized content, keeping spam at bay. It also collapses answers if the credibility of the answer or author is doubted or there are more than a certain number of downvotes.  We refrain from using Quora or other similar forums as they do not have a public corpora and the content posted on them is guarded by copyright\footnote{We contacted Quora, but they denied us permission to scrape data.}. 

Yahoo! Answers\footnote{https://answers.yahoo.com/} is a CQA Website, with all content open to browse and download. The Yahoo! Answers L6 dataset is provided by Yahoo! Research Alliance Webscope program\footnote{https://webscope.sandbox.yahoo.com/} on signing of a non-commercial research only agreement. The corpus consists of around 4.4 million  anonymized question threads and their corresponding answers. In addition to them , the corpus contains question specific metadata like the best answer, question category and question language. A best answer is mandatorily selected either by the asker, or by the users in the thread through upvoting, if the asker did not select a best answer.
We only deal with English language question threads. Figure \ref{fig:xml} is an instance of a Yahoo! Answers question thread.

We then process the dataset as follows:
\begin{enumerate}
\item \textbf{Removing low content question threads}: To maintain the quality of the CQA dataset, we filter out question threads with less than 5 answers. We also remove threads where the best answer is less than 100 words in length. We remove threads where answers other than the best answer together have less than 200 words. 
\item \textbf{Reference Generation}: We aim to have 100 word reference summaries. However many question threads have best answers with summaries longer than 100 words. In such cases, we choose important sentences/phrases and compress the best answer to less than hundred words to use it as reference. We follow a feature based approach to do the same. We iterate over every sentence in the best answer and assign a priority score to them. We look out for the proper nouns, common nouns, multiple consecutive delimiters (demarcating beginning of a new paragraph), presence of question words etc. to assign priority to a sentence. We take the weighted average of the features by assigning weight 1 to proper nouns, consecutive delimiters in the beginning and presence of question words in the sentence. We assign a weight of 0.25 to each common noun in the sentence\footnote{\url{https://medium.com/@acrosson/summarize-documents-using-tf-idf-bdee8f60b71}}. The linear combination of these weighted features forms the priority of the sentence. We sort sentences according to decreasing order of priority score. We pick these sorted sentences until 100 words are exceeded. After discarding the remaining sentences, we rearrange the selected sentences to the order present in the original document. If the reference exceeds 100 words, we truncate it after the hundredth word.
\item \textbf{Lack of information in the reference}: In some cases, there may be an opinion which is popular among candidate documents, that is not part of the best answer. Most summarization algorithms generate summaries based on popularity of a view point in candidate documents. In such case it would not be a fare reference, if a popular view point is not part of the ground-truth. In this case, we use results from existing CQA summarization research. 

According to an analysis by \citet{liu2008understanding}, on a portion of the same corpus, only $48\%$ of the question threads have a unique best answer. A unique best answer signifies that the best answer can independently summarize all other answers. We find that this number is higher in case of factual question threads and lower in case of opinion based questions threads. We transform the problem of validating our reference to the problem of determining if the best answer is indeed a unique best answer to the question thread. If that is not the case, we drop the question thread from our corpus. 

In corpora where there is not enough community engagement to use the number of upvotes for comparison, it is a task to find the best answer. We create bag of word entities corresponding to every answer including the best answer. We then create a fully connected graph with the bag of words representation of these answers as vertices. We compute the tf-idf scores corresponding to each vertex. We then assign edge weight to all edges, equal to cosine similarity between tf-idf vectors of the participating vertices. We then sum over all the outgoing edges at a vertex, iteratively for all vertices. We call these scores corresponding to a vertex, the `answer's cumulative correlation score'.  If we have a clear winning (maximum) cumulative correlation score, and that is the annotated best answer, we go forward with processing. If that is not the case, we drop the question thread from our corpus.

\item \textbf{Excess information in reference}: Sometimes there exists a view point in the reference summary that is not resonated by any of the other documents. In such cases, no extractive or abstractive summarization technique would be able to generate the reference summary. As a result, we iterate over the sentences of the reference and find its correlation to other documents. In case the correlation of more than 2 statements does not exceed 0.5 with any of the documents, we do not add the question thread to our corpus. In case there is a single sentence in the reference with low correlation to the other documents, we let it remain as no summarization technique would be able to generate it and all techniques would be scored equally.

\end{enumerate}
The CQASumm corpus generation workflow has been depicted in Figure \ref{fig:CQASummPipeline}.

\section{Experiments}
The Yahoo! Answers L6 dataset has more than 4.4 million question threads. Our filtering techniques exclude approximately $90\%$ of the dataset. Since there can be multiple versions of this dataset, using different filtering techniques, we do not mention a fixed size for the sampled dataset. Due to memory and processing constraints, we randomly choose 100 thousand question threads from the remaining dataset to conduct our experiments. All further results refer to this sampled subset of 100 thousand instances as our corpora. 
\subsection{CQASUMM Quality}
We evaluate the quality of our corpus. We conduct experiments to conclude the following:
\begin{itemize}
\item \textbf{Statistics}: The sampled corpus has $1,00,001$ question threads and $12,01,744$  answers. A question thread has an average of $12.02$ answers. An average answer in the corpus has 65.03 words. Similar to the DUC and TAC standard multi document summarization datasets, our reference summaries are 100 words long (see Table \ref{tab:corpus_stats}).

\item \textbf{Domain Distribution}: We study the domain distribution of our corpus using the category and sub category metadata that come along with the question thread. The question threads range over a wide variety of domains; however the most popular ones are ``Entertainment \& Music'' and ``Family and relationships''. Other fairly popular categories for questions in the corpora are ``Computers and Internet'', ``Beauty and Style'' and ``Games and Recreation''.
\item \textbf{Heterogenity}: Although the answers in the corpus are anonymized, we create a map of user ids to find the number of distinct authors. Our CQASumm corpus happens to be one of the most heterogeneous summarization corpus with over 1.2 million answers from mostly distinct authors, each author having a different writing style. 
\item \textbf{User Demographics}: According to other studies\footnote{https://www.quora.com/What-are-the-demographics-of-people-who-use-Yahoo-Answers} on the same corpus, majority of users on Yahoo! Answers are American and male. There is also a heavy population of trollers and spammers. Sarcasm content in the corpus is high. Users of the community particularly like to post polls: which are questions of the type ``A or B$?$'' 
\end{itemize}

\begin{table}
\caption{Statistics of the Yahoo! Answers L6 dataset after sampling for creating annotated summarization corpus.}
  \label{tab:corpus_stats}
  \begin{tabular}{c|c}
    \toprule
    Feature &   Frequency\\
    \midrule
    $\#$ Question Threads & 100001 \\
    $\#$ Answer documents  & 1201744 \\
    $\#$ Answers/thread  & 12.017 \\
    $\#$ Words/answer  & 65.027 \\
    $\#$ Words/reference  & 100 \\
    
  \bottomrule
\end{tabular}
  
%\vspace{-5mm}
\end{table}

\begin{table}
  \caption{Comparing performance of popular multi document summarization algorithms on 100 thousand  randomly sampled instances of CQASumm. We use ROUGE-1 and ROUGE-2 scores to evaluate summaries of 100 words.}
\label{tab:baselines}
  \begin{tabular}{c|c|l}
    \toprule
    Algorithm &  ROUGE-1 & ROUGE-2\\
    \midrule
     KL Divergence (Greedy) & 0.240 & 0.038 \\
     SumBasic & 0.277 & 0.041 \\
     LexRank (Graph Based)  & 0.284 & 0.047\\
     TextRank (Graph Based) & 0.278 & 0.048 \\
    
  \bottomrule
\end{tabular}
  \vspace{-5mm}
\end{table}

\subsection{Baselines}

After we generate the corpus, we evaluate the performance of well-known multi document summarization techniques on our dataset. We use popular summary evaluation metric ROUGE score for this purpose. We use ROUGE-1 and ROUGE-2 scores to compare baseline MDS summaries to our reference summaries. 
\begin{itemize}
\item \textbf{KL Divergence}: Introduced by \citet{haghighi2009exploring}, KullbackLeibler divergence is the measure of how one probability distribution is different from a second, reference probability distribution\footnote{\url{https://en.wikipedia.org/wiki/KullbackLeibler_divergence}}. The algorithm attempts to minimize the KL divergence between the candidate documents and the generated summary. 

\item \textbf{SumBasic}: One of the most popular baselines used in the summarization literature, SumBasic was introduced in \cite{nenkova2005impact} and refined in \cite{vanderwende2007beyond}. It is a frequency based summarization system and models the appearance of words in a summary as a multinomial distribution.

\item \textbf{Lexrank}: A graph based unsupervised learning algorithm introduced by \citet{erkan2004lexrank}. It is drawn upon the popular PageRank and HITS algorithms. A fully connected graph is built with sentences as vertices. Edge weights are assigned by computing cosine similarities between tf-idf entities of the two participating vertices. A threshold is decided; edges with weights below which are dropped. More the number of edges a vertex has, more important the corresponding sentence. 

\item \textbf{TextRank}:  It is also built on the PageRank algorithm \cite{page1999pagerank}. It tries to choose sentences such that the information disseminated is as close as possible to the original documents. The PageRank value of a webpage is the probability of a user opening that page. Similarly in TextRank, the score is the similarity of other sentences to that particular sentence.
\end{itemize}

\subsection{Comparative Analysis}
Table \ref{tab:baselines} shows the comparison. We find that LexRank performs the best (0.284) when using ROUGE-1 as the metric for evaluation while TextRank performs the best (0.048) when using ROUGE-2. Overall, all four baselines appear to be giving similar ROUGE scores. However on manual inspection of baseline summaries(see Table \ref{tab:AllTable}), we find that sentences chosen by them are different.

On upper bound evaluation (explained in Section \ref{RelatedWorks}) on a portion of our corpus, we are able to reach up to 0.49 ROUGE-1 score and 0.23 ROUGE-2 score extractively. Even higher scores can be obtained by choosing sentences abstractively. The baselines however, give a maximum of 0.284 ROUGE-1 and 0.048 ROUGE-2 extractively. We see there is a lot of scope for improvement. We hence introduce the task of CQA summarization.

\section{Opiniosumm Architecture}
We broadly classify non-factoid answers as factual or opinion based. Opinion based sentences might be any positive, negative or neutral. We find that highly polar opinionated sentences seldom exist in the reference. However, such sentences frequently appear in candidate documents . On inspecting the CQASumm dataset, the baseline summaries and their performance, we find that existing methods sometimes add highly polar opinion based sentences to summaries. In some cases, opinions on the same entity, belonging to different polarities were added to the same answer. Also highly polar sentences are found most likely to reduce ROUGE scores. 
We suggest an algorithm, called OpinionSumm to deal with such instances. 

Our corpus contains both opinion and fact based answers in approximately $65\%-35\%$ ratio. We  separate opinion based text from fact based ones for further processing. We find that often a single answer contains facts as well as opinion. We also find that opinions may change within an answer as the entity in context changes. As a result we first separate tiles\footnote{A tile is a multi paragraph segment of text that represents a single passage or sub topic \cite{hearst1997texttiling}.} in text.  We then classify tiles as fact based/opinionated and if opinionated, into positive/negative polarity.

\begin{figure*}
\begin{center}
\includegraphics[height=2in, width=\linewidth]{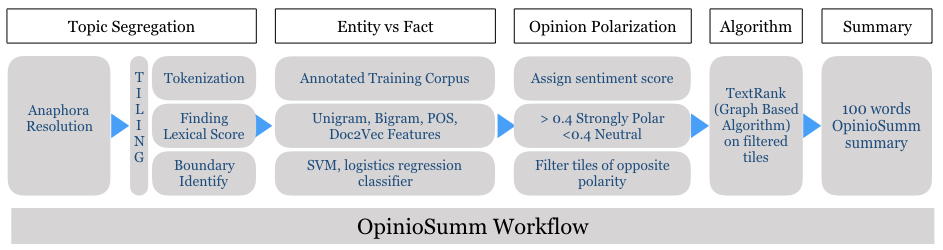}
\caption{We propose OpinioSumm: an opinion based summarization algorithm, built on top of TextRank to generate hundred words extractive summary in CQA. We resolve anaphora and segregate it  based on topic shifts. After tiling the text, we classify it as either fact or opinion. We next compute the polarity of question and answer tiles. We filter out tiles of polarity opposite to that of the question and pass the filtered tiles to TextRank for Summarization } \label{fig:OpinioSummPipeline}
\end{center}

\end{figure*}

We follow the steps described below (see Figure \ref{fig:OpinioSummPipeline}) to build Opinionsumm. 

\subsection{Topic Segregation}
\textbf{Anaphora Resolution}: We first resolve anaphora in the text. We use BART toolkit \cite{broscheit2010bart} on paragraph-based answers for anaphora resolution. We iteratively pass paragraphs in answer texts to this module. The REST-based web service of BART returns co-reference chains with each chain having a unique SetID attribute. We replace pronouns in the text with the nouns. This makes sentences within a paragraph independent candidates for the summary. This also allows us to follow entity-grid like approaches to detect entity changes, as discussed in \cite{steinberger2007two}

\textbf{Tokenisation}: Text tokens in the answers are identified. Any markup or image content is skipped, and remaining text is broken into tokens. Stop words are also removed. Inflected nouns and verbs are reduced to their roots by morphological analysis.
Next the text is divided into pseudo-sentences of length $w$ to make ground for uniform comparison between sentences of varying length. These pseudo sentences are referred to as token sequences. 
We create a dictionary of the token roots, and store the indices of the token sequences they occur in along with their respective counts. We also keep track of paragraph breaks within the text. 
 
 \textbf{Lexical Score Determination}: We use the vocabulary introduction method to compute lexical score detection. We assign a score to every token sequence gap based on the number of new words introduced in the interval considering the sequence gap as the mid point. The number of words which are never seen before in the  left token-sequence is added to the number of never seen before words in the token-sequence to the right. This number is divided by twice the token sequence length.
$score(i) = numNewTokens(b_1)+ numNewTokens(b_2)/w*2$, where $i$ is the token sequence gap index, $b_1$ and $b_2$ are the token sequences to the left and right of the $i$th gap and $w$ is the length of the token sequence.

 \textbf{Boundary Identification} : We assign a depth score to each token sequence gap as discussed above. The score stands for how strongly the cues of sub-topic change in sequences on either side of the sequence gap. Whenever the score exceeds a threshold, we create a new tile. 

\subsection{Entity vs Facts}
We classify each tile as fact or opinion.  Drawing inspiration from \cite{yu2003towards} and \cite{volkova2017separating} (both of which are popular works on separating facts from opinion), we train a SVM\footnote{We also tried Logistic Regression and vanilla Neural Networks; but they did not perform well.} model with a set of annotated documents. We build this annotated document set by taking factual instances from the Gigaword English news corpus. We use the twitter opinion dataset used by \citet{volkova2017separating} to collect opinionated documents. We use features such as unigrams, bigrams, trigrams and parts of speech. We also incorporate semantic information by embedding the the tiles in Doc2Vec space, trained on the Wikipedia English corpus. By the end of this module, we classify each tile as either fact or opinion based. We tabulate the performance of our fact/opinion classification algorithm at Table \ref{tab:factvsopinion}.  

\subsection{Opinion Polarization}
We use sentiment analysis to gauge the sentiment of the question. We compute the sentiment of the answer tiles classified above as opinion based. We use Serendio \cite{yadav2013serendio} to find opinion polarity at sentence level. The tool offers a normalized sentence sentiment score per
class (Positive and negative) using Sentiwordnet \cite{baccianella2010sentiwordnet} on feature words. We label the class with higher score, the winning class. We label a sentence strongly 'Positive' or 'Negative' if the dominant class score in the sentence is greater than 0.4. If the dominant class score is less than 0.4, we tag the sentence as neutral. An opinion based answer tile is assigned the same polarity as held by majority of constituting sentences.  

\subsection{Summary Generation}
We hypothesize:
\begin{enumerate}
\item If most of the tiles in a question thread are facts, the best answer is most likely to contain facts.
\item If a question statement has a strong polarity, most answer tiles would bear the same polarity.
\end{enumerate}  
As a result, for a neutrally opiniated question, we only pass fact based tiles and weakly opiniated tiles to TextRank for summary computation. If the question statement has strong polarity, we look into the constituent tiles for insight. In case most tiles have a strong polarity, we use a ternary classifier to segment the tiles according to their polarities (positive, negative and neutral). If there is a leading cluster (based on size) and the cluster is coherent, we pass only the leading cluster tiles to TextRank. If that is not the case, we feed all tiles to TextRank, as in the generic setting.

\subsection{Evaluation}
TextRank (a PageRank based algorithm, discussed in Section 4) returns a hundred word extractive summary. We use ROUGE-1 and ROUGE-2 score to compare the performance of generic TextRank vs OpinioSumm in Table \ref{tab:opiniosumm}. We show the summary produced by OpinioSumm beats TextRank (the best baseline) by $4.6\%$ on ROUGE-1 evaluation and $2.08\%$ on ROUGE-2 evaluation .

\begin{table}
 \caption{Classification of answer tiles into fact and opinion. We use unigrams, bigrams, trigrams and doc2vec  as features to train Logistic Regression and SVM.}
  \label{tab:factvsopinion}
  \begin{tabular}{ccl}
    \toprule
    Model &  F-score\\
    \midrule
     Logistic Regression & 0.59 \\
     SVM & 0.62 \\
  \bottomrule
\end{tabular}
\vspace{-5mm}
\end{table}
\vspace{1mm}
\begin{table}
\centering
\caption{Comparison of TextRank, and  OpinioSumm, on 0.1 million randomly sampled instances from CQASumm. We use ROUGE-1 and ROUGE-2 scores to evaluate summaries of 100 words.}\label{tab:opiniosumm}
  \begin{tabular}{ccl}
    \toprule
    Algorithm &  ROUGE-1 & ROUGE-2\\
    \midrule
     TextRank & 0.278 & 0.048 \\
	 OpinioSumm & 0.291 & 0.049 \\    
  \bottomrule
\end{tabular}
\vspace{-5mm}
\end{table}

\begin{table*}
\centering
\caption{An example CQA thread along with its six answer documents. Reference is the summary generated by CQASumm. We show the results of the four baselines: KL-Divergence, SumBasic, LexRank and TexRank on the question thread. Although the ROUGE-1 and ROUGE-2 scores in the different baselines are almost similar, we see that the sentences picked by them differ. }\label{tab:AllTable}
\scalebox{0.8}{
    \begin{tabular}{ | l | p{15cm} |}
    \hline
    Question & Why should I buy a IPOD? \\ \hline
   Best Answer & iPods are not just successful because they look good; they are a masterpiece of usability too. Other MP3 players come with software that isn't as good as iTunes, and you cannot use the best online store, iTunes Music Store. Sure, you get Napster and all the OD2 sites, but these are harder to use, often more expensive and more heavily DRMed.On batteries, they are much better since launch. If it dies within the first year, you can get Apple to pick it up and replace it for you within a week. The same is true for headphones and other accessories, though you don't have to send these back.Overall, Apple still offer the best deal. Of couse, wait til January and the price will drop. Rumours are also around that the next iPod to be unveiled at Macworld January will feature wireless song transfer and/or a camera. Wait and see. \\ \hline
    Answer 1 & You shouldnt! there are much better, cheaper options out there.\\ \hline
    
    Answer 2 & With the market saturation that Apple has with the ipods, it's easy to find help when needed. iPodLounge, iPodHacks, and such sites provide plenty of information on extending your iPod; i haven't found a fan-base like that for any other player.So, iPods are not superior players, but coupled with tech support, accessories, a large fan-base, and "projected longevity", ipods, in my opinion are good buys. And, if you buy iTunes music/video, the iPod is the only player on the block for you (unless you know how to crack DRM). \\ \hline
    
   Answer 3 & Bercause they are trendy. If all you want is a good mp3 player buy something else in my personal opinion. They do have a nice look, but are overpriced.\\ \hline 
   
   Answer 4 & You shouldn't. Sure they are nice looking and trendy, but keep in mind that the battery is not replaceable. When it dies in a year (can no longer be recharged) you will have to fork out the cash to buy another iPod.  There are better mp3 players out there, and way cheaper too. \\ \hline
   
   Answer 5 & I just bought the nhj personal video player it plays tv shows you record mp3's dvd movies you record to it you can store photos and it is the same size as an ipod you must lookt at it it is on sale for $\$189$ at geeks.com the sale ends soon go to their site fast see link below \\ \hline
   
   Answer 6 & If you remove the impulse response (it looks good or you dislike Apple products), and do not care about an integrated FM radio or built in recording, then the main reason is the integration between the iPod player with the PC/Mac software with the online store: The PC/Mac iTunes software is VERY easy to use and organize your music collection, whether mp3, mp4, AAC, or lossless. Features like the ability to "rate" your songs and create "Smart Playlists" (e.g. all  songs from the 80's that are 5 star plus the last 20 songs I recently listened to) are unique to iTunes/iPods (so far?). In addition, it will easily rip your CD's into the proper format for using on your PC and iPod. Also, if you are inclined to buy music over the internet, it is the easiest to use (although be cautioned that whatever you buy ONLY works on iPods or your PC/Mac). There are, or course, other comparisons and features to look at! \\ \hline
   
   Reference & Other MP3 players come with software that is n't as good as iTunes , and you can not use the best online store , iTunes Music Store. Sure , you get Napster and all the OD2 sites , but these are harder to use , often more expensive and more heavily DRMed. If it dies within the first year , you can get Apple to pick it up and replace it for you within a week. Overall , Apple still offer the best deal. Rumours are also around that the next iPod to be unveiled at Macworld January will feature wireless song transfer and/or a camera. \\ \hline 
    
   KL-Divergence & you shouldnt!there are much better, cheaper options out there.With the market saturation that Apple has with the ipods, it's easy to find help when needed.And, if you buy iTunes music/video, the iPod is the only player on the block for you (unless you know how to crack DRM).Bercause they are trendy.If all you want is a good mp3 player buy something else in my personal opinion.They do have a nice look, but are overpriced.You shouldn't.In addition, it will easily rip your CD's into the proper format for using on your PC \& iPod.Also, if you are inclined to buy music \\ \hline
   
    SumBasic & there are much better, cheaper options out there.With the market saturation that Apple has with the ipods, it's easy to find help when needed.So, iPods are not superior players, but coupled with tech support, accessories, a large fan-base, and "projected longevity", ipods, in my opinion are good buys.And, if you buy iTunes music/video, the iPod is the only player on the block for you (unless you know how to crack DRM).Bercause they are trendy.They do have a nice look, but are overpriced.You shouldn't.Features like the ability to "rate" your songs and create "Smart Playlists" (e.g. all songs from the 80's \\ \hline
    
     LexRank &  iPodLounge, iPodHacks, and such sites provide plenty of information on extending your iPod; i haven't found a fan-base like that for any other player.And, if you buy iTunes music/video, the iPod is the only player on the block for you (unless you know how to crack DRM).Bercause they are trendy.If all you want is a good mp3 player buy something else in my personal opinion.When it dies in a year (can no longer be recharged) you will have to fork out the cash to buy another iPod.There are better mp3 players out there, and way cheaper too.I just bought the \\ \hline
     
      TexRank & there are much better, cheaper options out there.With the market saturation that Apple has with the ipods, it's easy to find help when needed.And, if you buy iTunes music/video, the iPod is the only player on the block for you (unless you know how to crack DRM).Bercause they are trendy.Sure they are nice looking and trendy, but keep in mind that the battery is not replaceable.When it dies in a year (can no longer be recharged) you will have to fork out the cash to buy another iPod.There are better mp3 players out there, and way cheaper too.I just bought \\ \hline
      
       OpinioSumm &  With the market saturation that Apple has with the ipods, it's easy to find help when needed. So, iPods are not superior players, but coupled with tech support, accessories, a large fan-base, and "projected longevity", ipods, in my opinion are good buys. Bercause they are trendy. iPodLounge, iPodHacks, and such sites provide plenty of information on extending your iPod; i haven't found a fan-base like that for any other player. Sure , you get Napster and all the OD2 sites , but these are harder to use , often more expensive and more heavily DRMed. If it dies within the first year , you \\ \hline
       \end{tabular}}
\vspace{-8mm}

\end{table*}

%%%%%%%%%%%%%%%%%%%%%%%%%%%%%%%%%%%%%%%%%%%%%%%%%%%%%%%%%%%%%%%%%
%%%%COMPLETE BELOW%%%%%%%%%%%%%%%%
%%%%%%%%%%%%%%%%%%%%%%%%%%%%%%%%%%%%%%%%%%%%%%%%%%%%%%%%%%%%%%%%

\section{Related Work} \label{RelatedWorks}
\subsection{Summarization Datasets}
Summarization datasets can be of two types: extracts and abstracts. Extracts are reference summaries which have sentences picked from candidate documents. Abstracts are references created manually or using some heuristic e.g., best answer, headline of a news article etc. Extracts cannot be used for training or learning abstractive summaries. Hence it limits their scope.

Apart from the DUC and TAC task datasets, some other multi document summarization corpora have been released, each with a different objective. Most of them belong to the news domain and are low in heterogeneousness and variance. We discuss attempts outside the news domain. \citet{benikova2016bridging} use Deutscher Bildungsserver\footnote{http://www.bildungsserver.de} (DBS) to generate coherent extracts for educational articles. \citet{zopf2016next} use the Wikipedia corpora's existing summaries and following information nuggets to create a heterogeneous corpora which can be used for abstractive MDS. \citet{nakano2010construction}  collect web documents related to  query sentences and use them as candidate documents along with manual summaries to perform MDS. \citet{cao2016tgsum} use news articles linked to a tweet to generate an abstractive MDS corpus. They use the news articles as candidate documents and the tweet as a reference in corpus generation.

\subsection{Extractive Upper Bound Evaluation}
Given a set of documents and a reference summary, it is possible to determine the highest ROUGE score that can be obtained by extracting sentences from the documents. One way of finding the ROUGE upper bound summary is via Integer Linear Programming. The problem is referred as the Budgeted Maximum coverage problem (BMCP) by \citet{takamura2010learning}. The algorithm maximizes the total benefits of the words covered by selected
sentences. The model can be expressed using three expressions. It aims to maximize benefit by selecting sentences, constrained by the answer length. BMCP is however NP-Hard. We use a greedy solution to approximate ROUGE upper bounds.    

\section{Corpus Reconstruction}
The L6 corpus can be requested from \url{https://webscope.sandbox.yahoo.com/}. Our scripts and experiments can be found at \url{https://bitbucket.org/tanya14109/cqasumm}.
 We encourage the community to replicate our results.

\section{Discussion and Future Work}
We presented the first annotated corpus for Community Question Answering thread summarization and provided baseline summaries for the same. With this we introduce the task of {\em CQA thread summarization}. The proposed framework would enable CQA services to generate question thread summaries automatically  (e.g., Answer Wikis on Quora). With pre-trained language models and parallelization, the method can also be executed at run time, i.e., when the user opens a specific thread. The model presents multi sentence summaries of popular user opinion omitting spam, slurs, sarcasm and advertisement. 

We also compare the performance of various multi- document summarization techniques on our generated corpus.   The corpus can also be used to study generic multi document summarization where the only popular datasets are from the DUC and TAC tasks; which are considerably small in size and low in variance.
The dataset can be used to train  deep neural multi sentence abstractive summarization model for community question answering. At present, there is a dearth of such models in multi document summarization due to the costs associated with acquisition of annotated data. 

We proposed OpinioSumm, our multi document summarization algorithm. OpinioSumm specialises in separating facts and opinions in answers and using them according to the question characteristics. This is essential only in mixed datasets like CQA where there is a fair share of both factual and opinion based information. 

\section*{Acknowledgement}
This work was partially supported by the Ramanujan Faculty Fellowship,  Early Career Award (SERB, DST) and the Infosys Center for AI, IIIT-Delhi, India.

%We also look forward to reduce the latency and generate summaries for a single thread on run time so they can be deployed as browser extensions by CQA forums.  

\bibliographystyle{ACM-Reference-Format}
\bibliography{sample-bibliography}

\end{document}